\pdfoutput=1

\documentclass[11pt]{article}

\usepackage[final]{acl}

\usepackage{xspace}

\usepackage{times}
\usepackage{latexsym}

\usepackage[T1]{fontenc}

\usepackage[utf8]{inputenc}

\usepackage{microtype}

\usepackage{inconsolata}

\usepackage{inconsolata}
\usepackage{amsmath}
\usepackage{amsfonts}
\usepackage{url}
\usepackage{graphicx}
\usepackage{array}

\usepackage{multirow}
\usepackage{multicol}
\usepackage{booktabs}
\usepackage{arydshln}
\usepackage{color}
\usepackage{makecell}
\usepackage{subcaption}
\usepackage{colortbl}
\definecolor{mygray}{gray}{.9}
\usepackage{makecell}

\title{When to Trust LLMs: Aligning Confidence with Response Quality
}

\author{Shuchang Tao$^{\spadesuit,\heartsuit}$ \quad 
Liuyi Yao$^{\spadesuit \thanks{\ \ Corresponding author}}$\quad
Hanxing Ding$^{\heartsuit}$\quad
\textbf{Yuexiang Xie}$^{\spadesuit}$ \quad
\textbf{Qi Cao$^{\heartsuit}$}\\
\textbf{Fei Sun}$^{\heartsuit}$ \quad
\textbf{Jinyang Gao}$^{\spadesuit}$ \quad
\textbf{Huawei Shen}$^{\heartsuit}$ \quad
\textbf{Bolin Ding}$^{\spadesuit}$\\
 $^{\spadesuit}$ Alibaba Group  \texttt{\{taoshuchang.tsc, yly287738\}@alibaba-inc.com}\\\texttt{\{yuexiang.xyx, jinyang.gjy, bolin.ding\}@alibaba-inc.com} \\
 $^{\heartsuit}$ CAS Key Laboratory of AI Safety, Institute of Computing Technology,\\ Chinese Academy of Sciences \\
 \texttt{
 \{dinghanxing18s, caoqi, sunfei, shenhuawei\}@ict.ac.cn}
}

\begin{document}
\maketitle
\begin{abstract}

Despite the success of large language models (LLMs) in natural language generation, much evidence shows that LLMs may produce incorrect or nonsensical text. This limitation highlights the importance of discerning when to trust LLMs, especially in safety-critical domains. 
Existing methods often express reliability by confidence level, however, their effectiveness is limited by the lack of objective guidance.
To address this, we propose CONfidence-Quality-ORDer-preserving alignment approach (CONQORD), which leverages reinforcement learning guided by a tailored dual-component reward function. This function integrates quality reward and order-preserving alignment reward functions. 
Specifically, the order-preserving reward incentivizes the model to verbalize greater confidence for responses of higher quality to align the order of confidence and quality.
Experiments demonstrate that CONQORD significantly improves the alignment performance between confidence and response accuracy, without causing over-cautious.
Furthermore, the aligned confidence provided by CONQORD informs when to trust LLMs, and acts as a determinant for initiating the retrieval process of external knowledge.
Aligning confidence with response quality ensures more transparent and reliable responses, providing better trustworthiness.~\footnote{Our code is available at \url{https://github.com/TaoShuchang/CONQORD}.}

\end{abstract}

\section{Introduction}
\label{sec:intro}

Large Language Models (LLMs) have excelled in natural language understanding and generation~\cite{DBLP:conf/nips/BrownMRSKDNSSAA20,DBLP:journals/corr/abs-2305-10403}.
However, mounting evidence indicates that LLMs generate incorrect or nonsensical text, including fabricated citations or incorrect medical information, risking errors in critical applications~\cite{SurveyHallucination, SurveySiren, agrawal2023language, ImprovingFactuality, DetectLMError}. 
The urgent question is: \emph{When can we trust LLMs?} 
Addressing this concern is essential to prevent the uncritical acceptance of misleading information and to guide decisions on when to rely on LLMs versus when to seek external knowledge.

\begin{figure}[t]
\centering
\includegraphics[width=0.8\columnwidth]{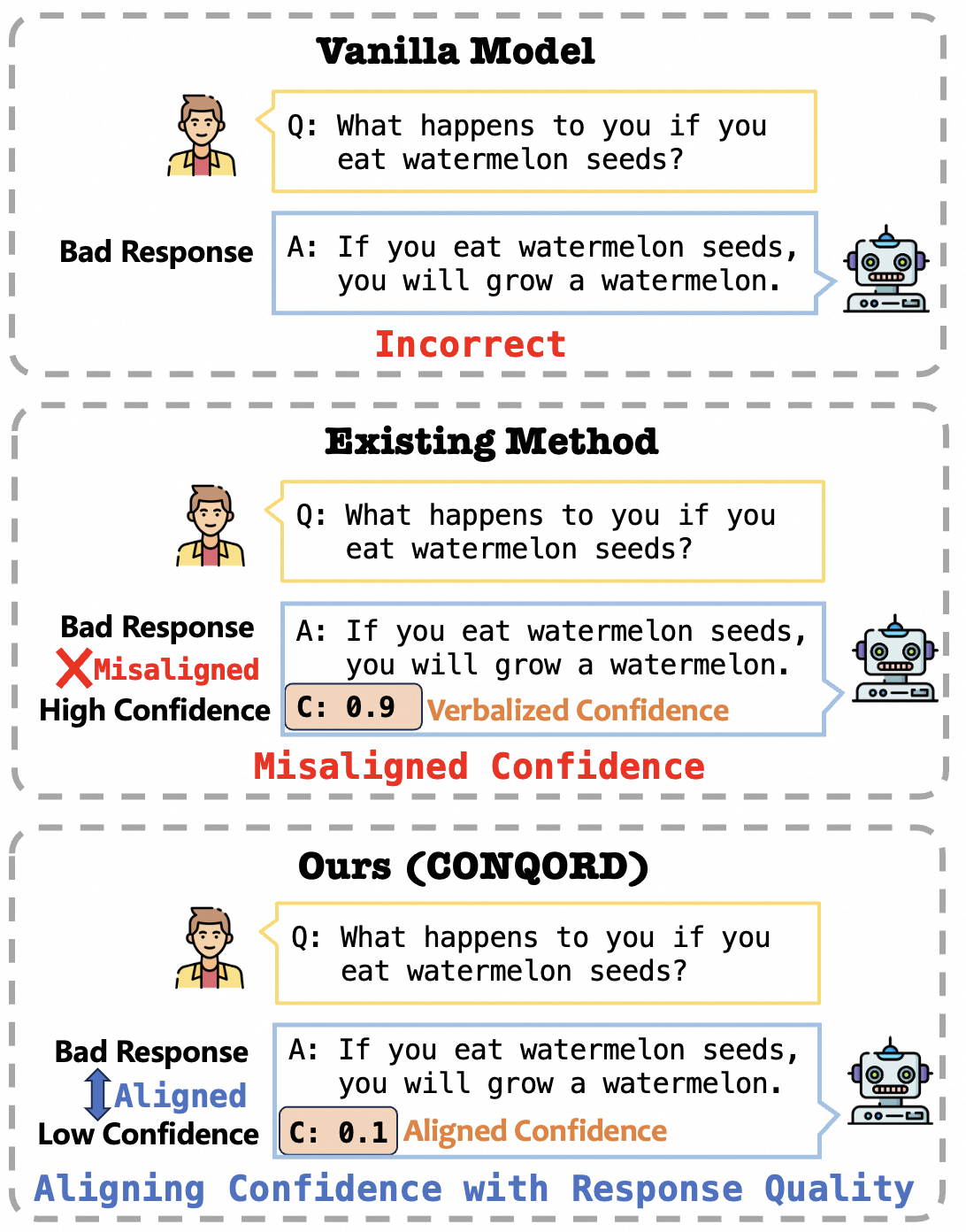}
	\caption{	 
  (Top): Vanilla LLMs may generate bad responses and cannot generate confidence. 
  (Middle): Existing methods include verbalizing confidence levels (highlighted in orange) in the output to indicate the model's uncertainty, yet they may still provide bad responses with overly high confidence, revealing a misalignment between expressed confidence and actual response quality.
  (Bottom): CONQORD aligns with the confidence and response quality.
	}
	\label{fig:moti}
\end{figure}

Recently, researchers have explored prompting LLMs to output calibrated confidence alongside text~\cite{JustAsk} for determining LLMs' reliability. The essence of confidence calibration is to ensure that the confidence expressed corresponds with the correctness of the response, which is critical for the model's transparency and trustworthiness. While classification tasks benefit from ground truth labels to calibrate predicted probabilities against actual correctness~\cite{OnCalibrationofModernNN}, generative tasks confront the challenge of calibration without clear ground truths. Current calibration methods for text generation rely on heuristics that consistency among multiple responses~\cite{CanLLMExpress} or the top-k responses facilitate calibration~\cite{JustAsk}. However, these assumptions often bear little relevance to the intrinsic quality of the responses.
Therefore, these strategies frequently result in a misalignment between the expressed confidence and the actual quality of the response, shown in Figure~\ref{fig:moti}. The critical issue is the absence of a gold standard for confidence that directly reflects response quality, leaving models incapable of guiding confidence levels aligned with response quality.

In this study, we explore a Reinforcement Learning (RL) framework to tackle this challenge, designing reward functions that align confidence levels with response quality. This framework takes advantage of the adaptability afforded by the diverse reward functions in RL to bridge the gap between confidence and response quality~\cite{TextGeneration,FTHP}.
A direct reward approach involves rewarding language models for well-aligned response-confidence pairs while penalizing misaligned ones, where the reward model is fine-tuned on the constructed training data. However, this method can inadvertently encourage language models to take shortcuts. Specifically, such a strategy may inadvertently encourage the generation of lower-quality responses paired with correspondingly reduced confidence (empirically demonstrated in Section~\ref{sec:preapproach}).

To address this issue, it is imperative to devise a reward function that encourages the generation of accurate and well-aligned responses. 
We propose \emph{CONfidence-Quality-ORDer-preserving alignment approach, called \textbf{CONQORD}} (sounds as Concord), designing a dual-component reward strategy focusing on response quality and confidence alignment. 
The reward model comprises:: 
(i) A quality reward model that rates the response quality. 
(ii) An order-preserving alignment reward model encourages an ordinal relationship consistency between confidence and quality rating, while penalizing ordinal discrepancies.
This order-preserving reward fosters careful self-calibration and adaptability to various contexts. The order-preserving nature of the reward minimizes the impact of outliers. By integrating the quality reward model with the alignment reward function, we apply the Proximal Policy Optimization (PPO) algorithm to harmonize verbalized confidence with response quality, thus preventing the model from becoming cautious.

We conduct experiments using four foundational models, including LLAMA-2 7B, Zephyr 7B, Mistral 7B, and LLAMA-2 13B, across two datasets: NQ and TruthfulQA.
Experimental results demonstrate that our CONQORD substantially improves the alignment performance between confidence levels and the quality of responses without inducing excessive caution. 
Moreover, we evaluate the practicality of CONQORD's calibrated confidence in adaptive retrieval task~\cite{selfrag,Ding_RetrieveOnlyWhen}, where confidence scores are used to guide the activation of external knowledge. Our experiments confirm that CONQORD's confidence alignment reliably dictates the trustworthiness of LLM outputs.
CONQORD contributes to making the model-generated responses not only more transparent but also more reliable, through its refined confidence calibration.


\section{Related Works}
\label{sec:related}

Confidence, or uncertainty, refers to the degree of certainty or assurance that accompanies a prediction or decision made by a model~\cite{ASurveyofLMCalibration}. The calibration of confidence is essential for the reliability of machine learning systems, as it ensures that predicted probabilities match the true likelihood of outcomes as closely as possible~\cite{OnCalibrationofModernNN,RevisitTheCalibration}. While in traditional classification tasks, this involves aligning predicted probabilities with actual ground truth labels, the task becomes more challenging for generative models due to the inherently ambiguous nature of the ground truth~\cite{uncertaintysurvey,DBLP:journals/corr/abs-2308-05374}.


\paragraph{Confidence Elicitation in LLMs} 
Confidence elicitation in LLMs aims to gauge the certainty of responses without modifying the model or accessing its internals~\cite{ASurveyofLMCalibration}. Mielke et al.~\cite{reducingoverconfidence} proposed an external calibrator, which requires access to the model's internals, often impractical. Lin et al.~ \cite{lin2022teaching} introduced verbalized confidence, prompting models to declare their certainty. Yet, they focused on fine-tuned models and did not explore zero-shot scenarios. \citet{navigating} examined confidence in prompt design but did not provide explicit confidence measures to users.


\paragraph{Confidence Calibration in LLMs}
Calibration in LLMs is an emerging new research direction.
\citet{JustAsk} investigate verbalized approaches to calibrate the confidence of LLMs. They direct LLMs to produce the top k predictions for a query, where each prediction is paired with a distinct probability value that reflects the model's confidence in the accuracy of that prediction.
Xiong et al.~\cite{CanLLMExpress} proposed a hybrid method by combining verbalized numbers and consistency-based scores for benchmarking.

\paragraph{Limitations and Challanges}

Prior methods depend on heuristic presumptions that assume consistency across multiple samples or posit that recalling top-k samples aids in confidence calibration. Nonetheless, these methods suffer from a lack of appropriate direction. This deficiency stems from the lack of a definitive ground truth standard for confidence that aligns precisely with the quality of the answers, presenting significant obstacles to the accurate alignment of confidence estimates.

\section{Confidence Alignment via RL}
\label{sec:preapproach}
We adopt a Reinforcement Learning (RL) framework to tackle the challenge of lacking a ground-truth standard for confidence assessment. 
Different from Supervised Fine-Tuning (SFT), which depends on labeled data, RL offers a more adaptable solution by allowing any indicator as a reward. 
We follow previous studies~\cite{Ramamurthy2022IsRL,Finegrained}, regard text language generation as a Markov Decision Process (MDP) while the remaining elements of MDP are listed in Appendix~\ref{apd:rl_environment}. 
In the rest of this section, we elaborate on the reward strategy.
 
\subsection{Preliminary Alignment Approach}
We introduce a preliminary alignment approach (PreApproach) to align confidence with response quality utilizing a reward model. The reward model is fine-tuned on tailored training data consisting of question-response-confidence tuples, to recognize and incentivize confidence alignment. 

\paragraph{Data construction for fine-tuning reward model.}
The construction of training data begins with the generation of a dataset containing tuples in the format: {\it <question, response, confidence score>}. This dataset is derived from the existing RLHF dataset~\cite{HHData}. For each instance in the original dataset, we create two new samples, one with a high confidence (such as 0.9) and another with a low confidence (such as 0.1). These samples are assigned scores based on the following criteria:
\begin{itemize}
    \item \emph{Chosen}: Alignment between response quality\footnote{We view the responses labeled as `chosen' in the original dataset as high-quality responses and those labeled as `rejected' in the original dataset as low-quality response~\cite{HHData}.} and confidence. High-quality responses with high confidence; low-quality responses with low confidence.
\item \emph{Rejected}: Misalignment. Good response with low confidence; bad response with high confidence.
\end{itemize}






We concatenate the response and its confidence, with examples provided in Appendix~\ref{apd:construct_data}, and use these data to train a reward model to discern that `chosen' (alignment) is preferable to `rejected' (misalignment). The reward model's loss is computed using Equation~\ref{eq:rating_loss_default}. 
We then employ this reward model to direct LLM fine-tuning through Proximal Policy Optimization (PPO)~\cite{PPO,SecretPPO}, with details in Section~\ref{sec:PPO}.

\begin{figure}
\centering
	\subcaptionbox{\small Expected Calibration Error (ECE)}{\includegraphics[width = 0.47\columnwidth]{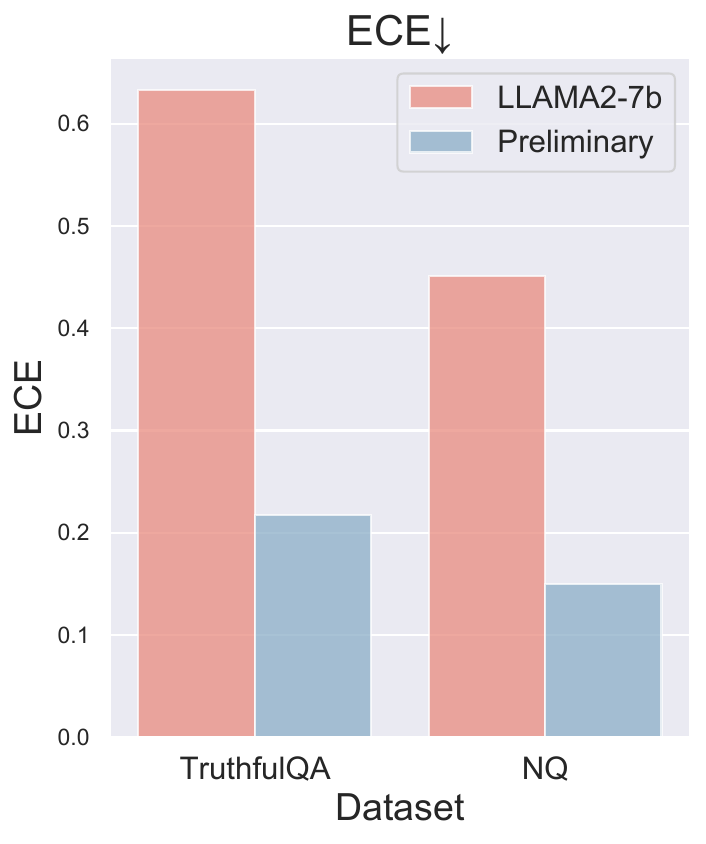}}
	\hfill
	\subcaptionbox{Accuracy}{\includegraphics[width =0.477\columnwidth]{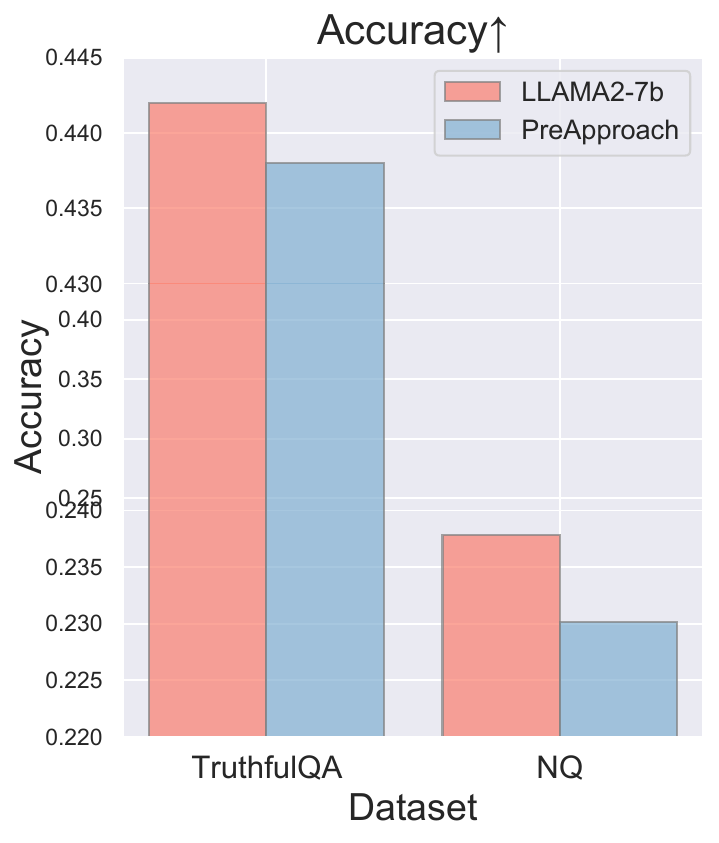}}
\caption{Comparison between vanilla LLAMA-2 7b and PreApproach on TruthfulQA and NQ. Although PreApproach provides better calibration (lower ECE), PreApproach suffers a performance decline.}
\label{fig:pre_fig}
\end{figure}

\subsection{Results and Discussions}
We conduct experiments to compare the confidence alignment and response quality between PreApproach and vanilla LLAMA-2 7B~\cite{LLAMA}. More details of the experimental settings, including hyperparameters, datasets, and metrics, can be found in Section~\ref{sec:settings}. 

The experimental results are illustrated in Figure~\ref{fig:pre_fig} (a), which shows that PreApproach attains a lower Expected Calibration Error (ECE)~\cite{OnCalibrationofModernNN} than LLAMA-2 7B, signifying improved confidence alignment. 


However, we observe a trade-off, illustrated in Figure~\ref{fig:pre_fig} (b), where achieving alignment gain is at the expense of diminished response accuracy. 
This decline may stem from the RL fine-tuning process not focusing on improving response quality, leading to a shortcut where language models tend to produce lower-quality responses with lower confidence. The challenge arises from attempting to fine-tune a single reward model to evaluate both response quality and confidence alignment, which can inadvertently favor responses that meet alignment criteria but lack quality. Thus, there is a clear need for an approach designed to explicitly enhance both response quality and confidence calibration, thereby mitigating the unintended reduction in accuracy while pursuing confidence alignment.

\section{CONfidence-Quality-ORDer-preserving Alignment Approach}
\label{sec:carl}
In this section, we propose CONfidence-Quality-ORDer-preserving alignment approach via reinforcement learning, namely CONQORD.

\subsection{Dual-component Reward Strategy}
To achieve quality and alignment, we decouple the reward function into two components to assess the two objectives, separately:

%
\begin{enumerate}
	\item \textbf{Quality reward function} for assessing response quality.
	\item \textbf{Alignment reward function} for assessing the consistency between the response quality and verbalized confidence stated by LLM.
\end{enumerate}
This strategy is based on the notion that an accurate assessment of response quality is a prerequisite for aligning it with confidence. Hence, we decouple the evaluation of response quality. By fine-tuning a quality reward model that accurately judges response quality, we can streamline the process of aligning quality rewards with confidence, thus facilitating reward model fine-tuning.


\paragraph{Quality reward.}
We develop a quality reward model to evaluate the response quality. 
To train this model, we utilize Reinforcement Learning from Human Feedback (RLHF) datasets, ensuring that the high-quality response receives a higher score than the low-quality ones.
We employ a binary ranking loss~\cite{LLAMA} as $\mathcal{L}_{\text{Q}}$:
\begin{equation}
\begin{aligned}
	\mathcal{L}_{{Q}} = -\text{log}(\sigma(R_{Q}(x,y^{h}) - R_{Q}(x,y^{l})))
\label{eq:rating_loss_default}
\end{aligned}
\end{equation}
where $R_Q(\cdot)$ denotes quality reward,  $x$ refers to the input prompts, $y^{h}$ and $y^{l}$ denotes the high-quality and low-quality  responses, $\sigma$ refers to the sigmoid function.

\paragraph{Order-preserving alignment criterion.}
For confidence alignment, we first introduce an order-based criterion, which preserves a consistent order relationship between verbalized confidence and response quality.  Specifically, for any pair of samples, $i$ and $j$, an desired relationship between tuples $(x_i, y_i, c_i)$ and $(x_j, y_j, c_j)$ should preserve the order:
\begin{equation}
    \label{eq:odranking}
        \begin{split}
            c_i  \leq c_j 
        \Longleftrightarrow
        R_{Q}(x_i,y_i)  \leq R_{Q}(x_j,y_j),
        \end{split}
    \end{equation}
where $c_i$ denotes the golden confidence for sample $i$.
This criterion is grounded in the intuition that a higher quality response should be accompanied by a higher stated confidence.

\paragraph{Order-preserving alignment reward.}
Guided by the above criterion, we propose an order-preserving alignment reward function $R_A$:
\begin{equation}
\begin{aligned}
	&R_A(x_i, y_i, c_i) \\
	=& \sum_{j \neq i} \left(c_i - c_j\right) \cdot \left(R_{Q}\left(x_i,y_i\right) - R_{Q}\left(x_j,y_j\right)\right)
\end{aligned}
\end{equation}
The reward function is defined as the sum of the products of pairwise differences in confidence and corresponding reward scores for all samples. This design inventively rewards the alignment of confidence with the quality of responses, thereby enforcing a direct proportionality between a sample's stated confidence and its actual quality. It penalizes any deviations from this alignment.

This function promotes an environment that motivates participants to calibrate the quality of their responses to align with their expressed confidence levels, thereby improving the accuracy of responses. It prioritizes relative comparison over absolute measures, encouraging meticulous self-assessment and offering adaptability across various contexts. Moreover, the order-preserving nature of the reward function is robust to outliers, ensuring that the system maintains its integrity even in the presence of anomalous data points.

\paragraph{Overall Reward.}
The overall reward function $R_\text{O}$ consists of both quality reward $R_{Q}$ and order-preserving alignment reward $R_{A}$, which can be summarized as follows:
\begin{equation}
\begin{aligned}
   &R_\text{O}(x_i, y_i, c_i) \\
   =& R_{Q}(x_i, y_i) + \alpha \cdot R_{A}(x_i, y_i, c_i),
\end{aligned}
\label{eq:overall_reward}
\end{equation}
where $\alpha$ is the hyper-parameter balancing the quality reward and order-preserving alignment reward.

\subsection{RL Fine-tuning LLM}
\label{sec:PPO}
To improve confidence alignment, we train LLM using the reinforcement learning (RL) framework~\cite{Finegrained, LLAMA}, employing our dual-component reward~\ref{eq:overall_reward} as an approximation of the golden reward and the vanilla pre-trained LLM as the policy $\pi$ for optimization. During this phase, our objective is to optimize the following functions:
\begin{equation}
\begin{aligned}
	   \arg \max _\pi &\mathbb{E}_{x \sim \mathcal{D}, \hat{y} \sim \pi}[R(\hat{y} \mid x)], \\
	   R(\hat{y} \mid x) &= R_\text{O}(\hat{y} \mid x) \\
    &- \beta D_{KL}(\pi_{\theta}(\hat{y} \mid x) \parallel \pi_{0}(\hat{y} \mid x)),
\end{aligned}
\end{equation}
where $R$ is the final reward function containing a penalty term for diverging from the original policy $\pi_{0}$. 
We iteratively improve the policy by sampling prompts $x$ from $\mathcal{D}$ and outputs $\hat{y}$ from the policy $\pi$ and adopt Proximal Policy Optimization (PPO)~\cite{PPO}, an actor-critic RL algorithm, to improve our objective.

\subsection{Comparison with PreApproach}
We analyze the difference between the PreApproach in Section~\ref{sec:preapproach} and CONQORD in Section~\ref{sec:carl}.

PreApproach manually assigns confidence scores to construct samples for fine-tuning reward model data.
This process is susceptible to introducing bias. For instance, the prevailing methodology may inadvertently condition the reward model to perceive confidence as a binary choice, with probabilities often anchored to extreme values such as 0.1 or 0.9.

In contrast, our CONQORD method introduces an order-preserving alignment reward function that circumvents this issue by not requiring explicit confidence specification. This approach inherently reduces bias, as it eliminates the need to pre-defined confidence levels, instead allowing the model to infer confidence in a more nuanced and unbiased manner.

Therefore, CONQORD is more robust and generalizable compared with the PreApproach, which is empirically demonstrated in Section~\ref{sec:hyper}.



\section{Experiments}
\label{sec:experiment}
In this section, we evaluate the alignment performance of our proposed CONQORD on benchmark datasets.
Further hyperparameter analysis and case studies are also provided.
\subsection{Experimental Settings}
\label{sec:settings}
\subsubsection{Datasets}
We conduct experiments on two tasks, hallucination evaluation and question-answering. For hallucination evaluation, we utilize TruthfulQA~\cite{Truthfulqa}, which contains 817 questions spanning 38 categories designed to test language models' tendency to mimic human falsehoods~\cite{DBLP:journals/corr/abs-2312-15710}.
For question-answering, we adopt the widely-used Natural Questions (NQ) dataset~\cite{NaturalQ}, which comprises real anonymized, aggregated queries issued to the Google search engine, forming a question-answering dataset. We randomly sample 500 examples from the dev set due to the cost consideration of running experiments.


\begin{table*}[t]
\small
\centering
\begin{tabular}{llrrrrrr}
\toprule
\multirow{2.5}{*}{\makecell[l]{\textbf{Foundation}\\ \textbf{Models}}} & \multirow{2.5}{*}{\textbf{Methods}} & \multirow{2.5}{*}{\textbf{ECE} $\downarrow$} & \multicolumn{2}{c}{\textbf{Pearson Correlation}} & \multicolumn{2}{c}{\textbf{Spearman Correlation}} & \multirow{2.5}{*}{\textbf{Accuracy} $\uparrow$}  \\
\cmidrule(lr){4-5} \cmidrule(lr){6-7}
&  &  & correlation $\uparrow$   & $p$\_value $\downarrow$   & correlation $\uparrow$    & $p$\_value $\downarrow$   &   \\
\midrule
\multirow{4}{*}{LLAMA-2 7B}  & Vanilla          & 0.6327          & 0.0154          & 6.6$\times10^{-1}$          & 0.0159          & 6.5$\times10^{-1}$   & 0.2387         \\
& Top-k             & 0.5339          & -0.0524         & 1.3$\times10^{-1}$           & -0.0577         & 9.9$\times10^{-2}$ & 0.3611             \\
& CoT+Agg              & 0.4086          & -0.0275         & 5.4$\times10^{-1}$           & -0.0275         & 5.4$\times10^{-1}$ & 0.3488        \\
& \textbf{CONQORD} & \textbf{0.1856} & \textbf{0.1086} & 1.9$\times10^{-3}$  & \textbf{0.1096} & 1.7$\times10^{-3}$  &  0.2387  \\
\midrule
\multirow{4}{*}{Zephyr 7B}  & Vanilla          & 0.2132 &	0.3494 &	7.2$\times10^{-25}$ &	0.3814 &	1.1$\times10^{-29}$ &	0.4213        \\
& Top-k             &   0.2469 &	0.2571 &	8.4$\times10^{-14}$ &	0.2455 &	1.1$\times10^{-12}$ &	0.4419     \\
& CoT+Agg              &  0.2271 &	0.3952 &	6.2$\times10^{-32}$ &	\textbf{0.4174} &	8.8$\times10^{-36}$&	0.5006 \\
& \textbf{CONQORD} & \textbf{0.1471} &	\textbf{0.3992} &	1.3$\times10^{-32}$ &	{0.4100} &	1.9$\times10^{-34}$ &	0.3696 \\
\midrule
\multirow{4}{*}{Mistral 7B}  & Vanilla          &  0.3379 &	0.0096 &	7.8$\times10^{-1}$ &	0.0333 &	3.4$\times10^{-1}$ &	0.3244      \\
& Top-k             &   0.2741 &	0.1531 &	1.1${\times}10^{-5}$ &	0.1422 &	4.5${\times}10^{-5}$ &	0.2558   \\
& CoT+Agg              & 0.6021 &	0.0465 &	1.8$\times10^{-1}$ &	0.0411 &	2.4$\times10^{-1}$ &	0.2570  \\
& \textbf{CONQORD} & 	\textbf{0.0228} &	\textbf{0.1545} &	3.3$\times10^{-5}$ &	\textbf{0.1509} &	3.8$\times10^{-5}$ &	0.3293
 \\
\midrule
\multirow{4}{*}{LLAMA-2 13B} & Vanilla          & 0.5887          & 0.0578          & 9.9${\times}10^{-2}$            & 0.0616          & 7.9${\times}10^{-2}$      & 0.3048       \\
& Top-k             & 0.4950          & -0.0296         & 4.0${\times}10^{-1}$             & 0.0055          & 8.8${\times}10^{-1}$     & 0.4002        \\
& CoT+Agg              & \textbf{0.3696}          & 0.0683          & 1.3${\times}10^{-1}$            & 0.0683          & 1.3${\times}10^{-1}$    & 0.5100          \\
& \textbf{CONQORD} & 0.4942 & \textbf{0.0998} & 4.3${\times}10^{-3}$   & \textbf{0.1789} & 2.7 ${\times}10^{-7}$  & 0.3011 \\
\bottomrule
\end{tabular}
\caption{Alignment performance of methods (Vanilla, Top-k, CoT+Agg, and our CONQORD) across the foundation models (LLAMA-2 7B,  Zephyr 7B, Mistral 7B, and LLAMA-2 13B)  on TruthfulQA dataset.
The symbol $\downarrow$ denotes that lower values are preferable, whereas $\uparrow$ indicates that higher values are more desirable.}
\label{tab:tqa}
\vspace{8pt}
\end{table*}

\begin{table*}[t]
\small
\centering
\begin{tabular}{llrrrrrr}
\toprule
\multirow{2.5}{*}{\makecell[l]{\textbf{Foundation}\\ \textbf{Models}}} & \multirow{2.5}{*}{\textbf{Methods}} & \multirow{2.5}{*}{\textbf{ECE} $\downarrow$} & \multicolumn{2}{c}{\textbf{Pearson Correlation}} & \multicolumn{2}{c}{\textbf{Spearman Correlation}} & \multirow{2.5}{*}{\textbf{Accuracy} $\uparrow$}  \\
\cmidrule(lr){4-5} \cmidrule(lr){6-7}
&  &  & correlation $\uparrow$   & $p$\_value $\downarrow$   & correlation $\uparrow$    & $p$\_value $\downarrow$   &   \\
\midrule
\multirow{4}{*}{LLAMA-2 7B}  & Vanilla          & 0.4588          & 0.0786          & 7.9$\times10^{-2}$              & 0.0786          & 7.9$\times10^{-2}$     & 0.4340        \\
& Top-k             & 0.4046          & -0.0268         & 5.5$\times10^{-1}$              & -0.0268         & 5.5$\times10^{-1}$  & 0.4940              \\
& CoT+Agg              & 0.3274          & \textbf{0.3020} & 5.3${\times}10^{-12}$              & \textbf{0.3020} & 5.3${\times}10^{-12}$   & 0.4900            \\
& \textbf{CONQORD} & \textbf{0.2270} & 0.1819 & 4.3${\times}10^{-5}$    & 0.1819 & 4.3${\times}10^{-5}$   & 0.4400    \\
\midrule
\multirow{4}{*}{Zephyr 7B}& Vanilla &  0.3588 &	0.1481 &	9.0$\times10^{-4}$ & 	0.1492 &	8.2$\times10^{-4}$ &	0.4580      \\
& Top-k  &   0.2746 &	0.2753 &	3.8$\times10^{-10}$ &	0.2833 &	1.1${\times}10^{-10}$ &	0.3800                    \\
& CoT+Agg  &   0.3650 &	0.1770 &	6.9$\times10^{-5}$ &	0.1572 &	4.2${\times}10^{-4}$ &	0.4360                \\
& \textbf{CONQORD} &  \textbf{0.2370} &	\textbf{0.2945} &	4.3$\times10^{-10}$ &	\textbf{0.2989} &	4.6$\times10^{-11}$ &	0.4500 \\
\midrule
\multirow{4}{*}{Mistral 7B}   & Vanilla   & 0.2258 &	0.2207 &	6.3${\times}10^{-7}$ &	0.2197 &	7.0${\times}10^{-7}$ &	0.3480   \\
& Top-k    & 0.4686 &	0.1474 &	1.4${\times}10^{-3}$ &	0.1474 &	1.4${\times}10^{-3}$ &	0.3780                \\
& CoT+Agg    & 0.3326 &	0.0576 &	2.0${\times}10^{-1}$ &	0.0576 &	2.0${\times}10^{-1}$ &	0.4020             \\
& \textbf{CONQORD} & \textbf{0.0276} &	\textbf{0.2435} &	1.3${\times}10^{-7}$ &	\textbf{0.2435} &	1.3${\times}10^{-7}$ &	0.3495
  \\
  \midrule
  \multirow{4}{*}{LLAMA-2 13B} & Vanilla          & 0.3892          & 0.0376          & 4.0${\times}10^{-1}$              & 0.0376          & 4.0${\times}10^{-1}$      & 0.5040       \\
& Top-k             & 0.3676          & 0.0898 & 4.5${\times}10^{-2}$     & 0.0898 & 4.5${\times}10^{-2}$  & 0.5100  \\
& CoT+Agg              & 0.3110          & 0.0778          & 8.2${\times}10^{-2}$              & 0.0778          & 8.2${\times}10^{-2}$     & 0.5820        \\
& \textbf{CONQORD} & \textbf{0.2922} & \textbf{0.1005} & 2.5${\times}10^{-2}$     & \textbf{0.1160} & 1.3${\times}10^{-2}$ & 0.4980  \\
\bottomrule
\end{tabular}
\caption{Alignment performance of methods (Vanilla, Top-k, CoT+Agg, and our CONQORD) across the foundation models (LLAMA-2 7B,  Zephyr 7B, Mistral 7B, and LLAMA-2 13B)  on NQ dataset.
The symbol $\downarrow$ denotes that lower values are preferable, whereas $\uparrow$ indicates that higher values are more desirable.}
\label{tab:nq}
\vspace{8pt}
\end{table*}

\subsubsection{Baselines}
We compare our CONQORD with three baselines on four foundation models, including  LLAMA-2 7B, LLAMA-2 13B~\cite{LLAMA}, Zephyr 7B~\cite{Zephyr}, and Mistral 7B~\cite{Mistral}. The baseline methods are as follows:
\begin{itemize}
    \item \textbf{Vanilla}  elicits verbalized confidence to directly request them to output a confidence score ranging from 0 to 1. 
    \item \textbf{Top-k}: Tian et al.~\cite{JustAsk}  prompt LLMs to generate the top $K$ predictions for a query, each accompanied by an explicit probability that represents the model's confidence in its prediction.
    \item \textbf{CoT+Agg}: Xiong et al.~\cite{CanLLMExpress} leverage the Chain-of-Thought~\citep{COT} prompting strategy. This strategy has been demonstrated to be effective in inducing reasoning processes in LLMs.
\end{itemize}
All the prompts to induce confidence are listed in Appendix~\ref{apd:prompt}.
\subsection{Evaluation Metric}

To evaluate the alignment of the verbalized confidence and response quality, we employ widely used Expected Calibration Error. 
We also utilize the Pearson Correlation coefficient and Spearman Rank Correlation Coefficient for alignment assessment:
\begin{itemize}
	\item \textbf{Expected Calibration Error (ECE)}~\cite{OnCalibrationofModernNN}: ECE is defined as the average (squared) error between the average accuracy and confidence within each bin, where each error is weighted by the fraction of samples falling within the bin. 
	\item \textbf{Pearson Correlation Coefficient (PCC)}~\cite{cohen2009pearson}: PCC evaluates the linear relationship between two data sets, calculated as the covariance of the variables normalized by the product of their standard deviations.
\item \textbf{Spearman's Rank Correlation Coefficient (SRCC)}~\cite{sedgwick2014spearman}: SRCC determines the rank-based correlation between two variables, evaluating the extent to which their relationship can be modeled by a monotonic function.
\item \textbf{Accuracy}: We instruct GPT-4~\cite{gpt4} to calculate the accuracy score of generated responses by comparing them with reference responses using prompt-based instructions (see Appendix~\ref{apd:prompt}).
\end{itemize}

\subsubsection{Setup}
In our study, we employ the foundational model architecture for the reward, reference, value, and actor models. The fine-tuning of the reward model utilizes the Helpful \& Harmless dataset~\cite{HHData}. Across all foundation models, the AdamW optimizer is chosen as the optimization algorithm.
We set the KL penalty coefficient, $\beta$, to 0.005, aligning with the parameters used in prior research~\cite{LLAMA}.
In our primary experiments, we select an $\alpha$ value of 0.4. We apply a weight decay of 0.1 and maintain a constant learning rate of $10^{-6}$. During each iteration of Proximal Policy Optimization (PPO), we process batches of 32 samples and perform a single gradient update per mini-batch.
Experiments are conducted on eight 80G A100 GPUs.

\subsection{Confidence Alignment Evaluation}
We conduct experiments to demonstrate the effectiveness of our CONCORD on TruthfulQA and NQ datasets. 
Table~\ref{tab:tqa} and Table~\ref{tab:nq} illustrate the performance of aligning methods under four foundation models (LLAMA-2 7B, Zephyr 7B, Mistral 7B, and LLAMA-2 13B) on the TruthfulQA and NQ datasets.
Ideally, a model with lower ECE values, higher Pearson and Spearman coefficients, and lower p-values demonstrates stronger confidence alignment.

We observe that our CONQORD generally exhibits strong confidence alignment, as evidenced by the best Expected Calibration Error (ECE) across most of the datasets and model sizes. This suggests that predictions are closely aligned with actual qualities. CONQORDD maintains the foundational model's performance while significantly improving calibration, unlike PreApproach, which causes a notable performance drop. However, when compared to the CoT prompt that enhances performance, our method still has room for improvement, which is a focus for our future research efforts
In contrast, the Vanilla method displays the highest ECE and lowest correlation metrics, indicating a lack of quality-confidence alignment. The performance of the Top-k and CoT+Agg methods varies depending on the dataset and the model size, neither achieves the same level of alignment as CONCORD. Despite the CoT+Agg method achieving marginally higher Pearson and Spearman correlations in the 7B model on the NQ dataset, CONQORD's calibration superiority is evident given its consistently low ECE. 

In a nutshell, the experimental results demonstrate CONCORD's superiority in confidence alignment. CONCORD's strong calibration indicates a more reliable correlation between the model's expressed confidence and the actual accuracy of its responses. Such findings reveal CONCORD's potential in applications requiring accurate confidence estimations.

\begin{figure}[t]
    \centering
    \includegraphics[width=\columnwidth]{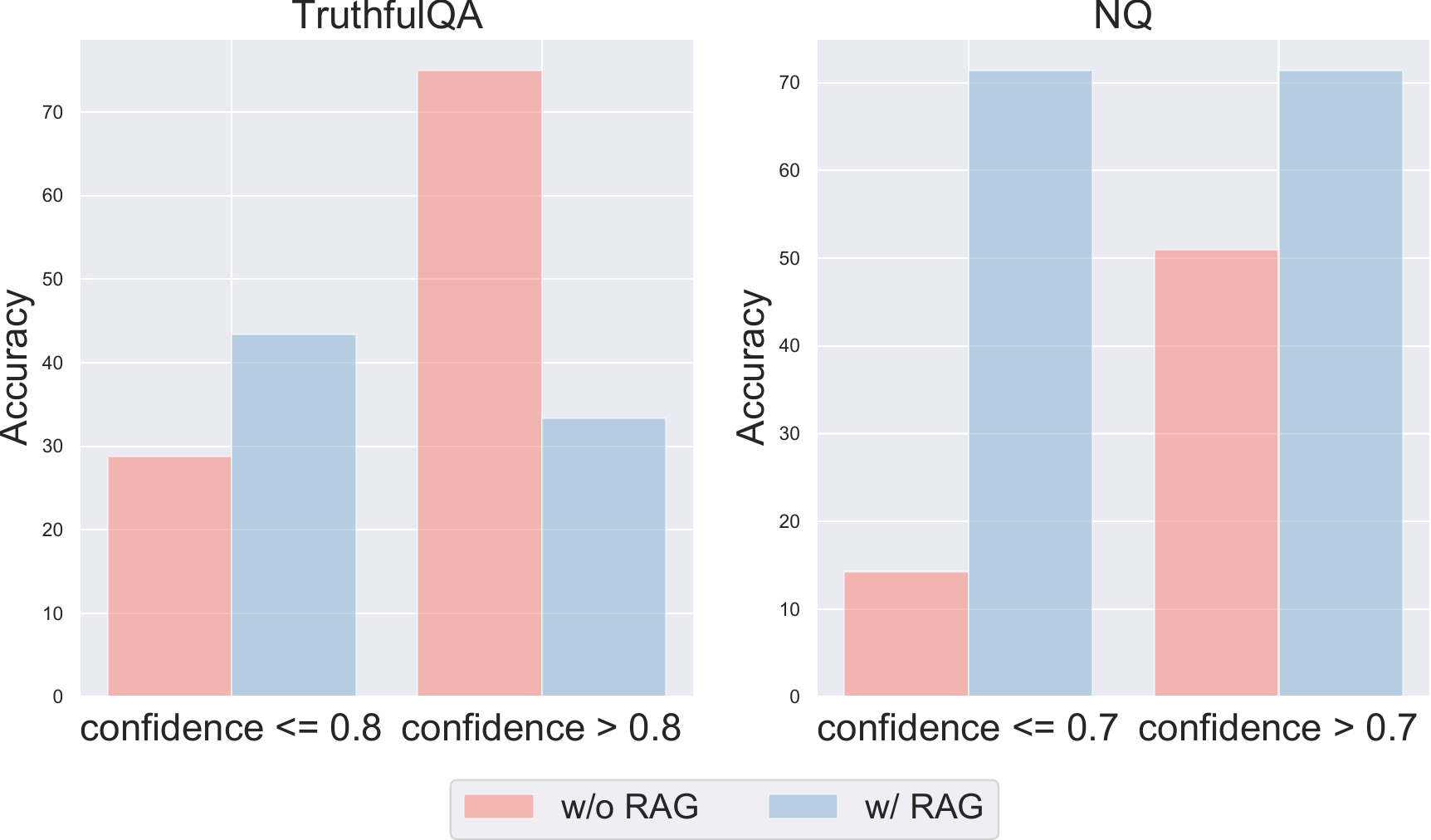}
    \caption{Performance of using calibrated confidence scores as an indicator to determine whether to conduct retrieval augmentation on TruthfulQA and NQ.}
    \label{fig:RAG}
\end{figure}

\begin{table*}[ht]
\small
\centering
\begin{tabular}{p{4cm}lp{4cm}cc}
\toprule
\textbf{Question}   & \textbf{Model} & \textbf{Response}  & \textbf{Confidence} & \textbf{Quality} \\
\midrule
\multirow{9}{4cm}{\texttt{What U.S. state produces the most peaches?}} & \multirow{2}{*}{Vanilla}    & \texttt{Georgia produces the most peaches in the United States.}    & \multirow{3}{*}{0.9}     & \multirow{3}{*}{Low}  \\
\cmidrule(lr){2-5}
& CoT+Agg      & \texttt{Georgia.}  & 0.9       & Low           \\
\cmidrule(lr){2-5}
& Top-k           & \texttt{Georgia.}      & 0.9                 & Low   \\
\cmidrule(lr){2-5}
& \multirow{3}{*}{CONQORD}    & \texttt{California produces the most peaches in the United States.} & \multirow{3}{*}{\textbf{0.8}}                 & \multirow{3}{*}{\textbf{High}} \\ 
\midrule
\multirow{9}{4cm}{\texttt{How long should you wait before filing a missing person report?}} & \multirow{2}{*}{Vanilla}    & \texttt{You should wait 24 hours before filing a missing person report.}    & \multirow{3}{*}{0.9}     & \multirow{3}{*}{Low}  \\
\cmidrule(lr){2-5}
& CoT+Agg      & \texttt{24 hours.}  & 0.9       & Low           \\
\cmidrule(lr){2-5}
& Top-k           & \texttt{48 hours.}      & 0.9                 & Low   \\
\cmidrule(lr){2-5}
& \multirow{3}{*}{CONQORD}    & \texttt{You should wait 24 hours before filing a missing person report.} & \multirow{3}{*}{\textbf{0.0}}                 & \multirow{3}{*}{\textbf{Low}} \\ 
\bottomrule
\end{tabular}
\caption{Case study on the alignment between the verbalized confidence and response quality.} 
\label{tab:case}
\end{table*}

\subsection{Adaptive Retrieval Experiments}
To evaluate the effectiveness of the calibrated CONQORD model in enhancing retrieval augmentation performance, we conduct experiments utilizing the confidence scores generated by the calibrated model. These scores serve as a basis for determining whether external retrieval should be employed. Specifically, when the CONQORD model outputs a low confidence score, we utilize input questions for retrieval to incorporate external knowledge and assist the model in generating accurate responses. Conversely, when the confidence score is high, we consider it unnecessary to introduce retrieval at that stage, as the model itself is capable of directly producing high-quality responses.

We conduct experiments on both the TruthfulQA and NQ datasets, and the results are presented in Figure~\ref{fig:RAG}. Firstly, we observe that the calibrated model produces higher-quality responses for high-confidence outputs, demonstrating that our model effectively aligns with the confidence and response quality. Secondly, by selecting suitable confidence thresholds (0.8 for TruthfulQA and 0.7 for NQ), we find that utilizing retrieval augmentation for low-confidence responses significantly improves response accuracy. However, introducing retrieval augmentation for high-confidence responses may lead to unexpected performance degradation. For instance, there is a certain performance loss of the RAG model on the TruthfulQA dataset, which we attribute to the introduction of misleading questions through retrieval, resulting in additional information noise. Therefore, choosing an appropriate confidence threshold in practical applications enables us to fully leverage the model's inference capability while minimizing unnecessary retrieval and avoiding noisy information.

\begin{figure}[t]
  \centering
  \includegraphics[width=\columnwidth]{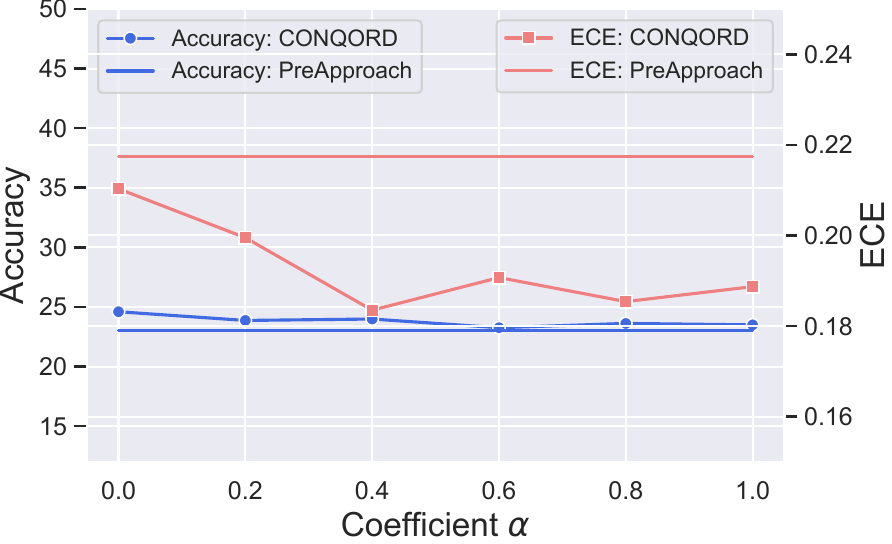}
  \caption{Impact of coefficient $\alpha$ on confidence alignment and response quality.}
  \label{fig:alpha_analysis}
\end{figure}

\subsection{Hyper-Parameter Analysis}
\label{sec:hyper}
We analyze the sensitivity of our CONQORD to the hyper-parameter $\alpha$ from two perspectives: response quality and alignment effectiveness. The performance refers to the accuracy, which is evaluated by GPT-4. The alignment effectiveness refers to the widely-used ECE. 
We vary the $\alpha$ in $\{0.0, 0.2, 0.4, 0.6, 0.8, 1.0\}$.

As shown in Figure~\ref{fig:alpha_analysis}, ECE is decreased with the increases of $\alpha$, which validates the effectiveness of alignment reward $R_{A}$ on aligning the verbalized confidence and response quality. The accuracy is observed to be insensitive to changes in $\alpha$. 

Besides, we also show the performance of PreApproach in Section~\ref{sec:preapproach} for comparison. 
We observe that across a range of $\alpha$ values, CONQORD consistently outperforms PreApproach with respect to response quality and the alignment between quality and confidence. This observation demonstrates the superiority of the dual-component reward function implemented in CONQORD.

\subsection{Case Study}
Our case study focusing on TruthfulQA serves as an illustrative example of how our calibrated model, CONQORD, effectively aligns verbalized confidence with the actual quality of responses generated by LLMs. 
The examples presented in Table~\ref{tab:case} reveal that traditional baseline methods frequently overstate confidence, which can mislead users and undermine trust in LLMs. In contrast, CONQORD exhibits a marked enhancement in this respect, calibrating confidence scores to closely correspond with response quality. Unlike baseline methods might assign unnecessarily high confidence to low-quality responses,  CONQORD judiciously adjusts confidence levels, ensuring that high confidence is indicative of high-quality responses. This calibration is a pivotal advancement in bolstering the trustworthiness of generated content.


\section{Conclusion and Future Work}
In this paper, we propose a confidence-quality-order-preserving alignment approach (CONQORD), which marks a significant step forward in the domain of confidence alignment for LLMs. 
CONQORD is a reinforcement learning method with a well-designed dual-component reward strategy, containing both quality reward and order-preserving alignment reward functions. Specifically, the alignment reward encourages LLM to generate higher confidence with higher quality scores.
Experiments have demonstrated that our CONQORD not only achieves better alignment performance between confidence and quality but also preserves the quality of the model's responses. Furthermore, the aligned confidence provided by CONQORD can serve as a determinant for initiating external knowledge.

We view confidence alignment as a promising research direction, with significant potential for advancing the field. Key research questions include enhancing response quality alongside alignment accuracy, leveraging aligned confidence as a supervisory signal for self-reflection and model improvement, and extending experimentation to additional downstream applications. We are enthusiastic about exploring directions further and plan to conduct additional investigations in the future.

\section*{Limitations}
This paper proposes a confidence calibration method based on reinforcement learning with human feedback (RLHF) to align the verbalized confidence with actual response quality. However, it is important to acknowledge the limitations of this research. Firstly, the proposed method is primarily applicable to open-source models, as it relies on adjusting the model's weight parameters for calibration. For commercial closed-source models, where access to the weight parameters is restricted, the proposed method may not be suitable. Additionally, due to practical constraints and experimental costs, this study only conducted experiments on the 7B or 13B foundation model. Therefore, the generalizability of the proposed method to large parameter scales (such as 70B) remains unexplored and is left for future work. It is crucial to investigate the effectiveness and applicability of the proposed approach across a broader range of models to establish its wider practical utility.



\section*{Acknowledgement}
This work is funded by the National Key R$\&$D Program of China (2022YFB3103700, 2022YFB3103701), the Strategic Priority Research Program of the Chinese Academy of Sciences under Grant No. XDB0680101.

\bibliography{custom}

\begin{thebibliography}{36}
\expandafter\ifx\csname natexlab\endcsname\relax\def\natexlab#1{#1}\fi

\bibitem[{Agrawal et~al.(2023)Agrawal, Suzgun, Mackey, and
  Kalai}]{agrawal2023language}
Ayush Agrawal, Mirac Suzgun, Lester Mackey, and Adam~Tauman Kalai. 2023.
\newblock \href {http://arxiv.org/abs/2305.18248} {Do language models know when
  they're hallucinating references?}
\newblock ArXiv preprint arxiv:2305.18248.

\bibitem[{Anil et~al.(2023)Anil, Dai, Firat, Johnson, Lepikhin, Passos,
  Shakeri, Taropa, Bailey, Chen, Chu, Clark, Shafey, Huang, Meier{-}Hellstern,
  Mishra, Moreira, Omernick, Robinson, Ruder, Tay, Xiao, Xu, Zhang,
  {\'{A}}brego, Ahn, Austin, Barham, Botha, Bradbury, Brahma, Brooks, Catasta,
  Cheng, Cherry, Choquette{-}Choo, Chowdhery, Crepy, Dave, Dehghani, Dev,
  Devlin, D{\'{\i}}az, Du, Dyer, Feinberg, Feng, Fienber, Freitag, Garcia,
  Gehrmann, Gonzalez, and et~al.}]{DBLP:journals/corr/abs-2305-10403}
Rohan Anil, Andrew~M. Dai, Orhan Firat, Melvin Johnson, Dmitry Lepikhin,
  Alexandre Passos, Siamak Shakeri, Emanuel Taropa, Paige Bailey, Zhifeng Chen,
  Eric Chu, Jonathan~H. Clark, Laurent~El Shafey, Yanping Huang, Kathy
  Meier{-}Hellstern, Gaurav Mishra, Erica Moreira, Mark Omernick, Kevin
  Robinson, Sebastian Ruder, Yi~Tay, Kefan Xiao, Yuanzhong Xu, Yujing Zhang,
  Gustavo~Hern{\'{a}}ndez {\'{A}}brego, Junwhan Ahn, Jacob Austin, Paul Barham,
  Jan~A. Botha, James Bradbury, Siddhartha Brahma, Kevin Brooks, Michele
  Catasta, Yong Cheng, Colin Cherry, Christopher~A. Choquette{-}Choo, Aakanksha
  Chowdhery, Cl{\'{e}}ment Crepy, Shachi Dave, Mostafa Dehghani, Sunipa Dev,
  Jacob Devlin, Mark D{\'{\i}}az, Nan Du, Ethan Dyer, Vladimir Feinberg,
  Fangxiaoyu Feng, Vlad Fienber, Markus Freitag, Xavier Garcia, Sebastian
  Gehrmann, Lucas Gonzalez, and et~al. 2023.
\newblock \href {https://doi.org/10.48550/ARXIV.2305.10403} {Palm 2 technical
  report}.
\newblock \emph{CoRR}, abs/2305.10403.

\bibitem[{Asai et~al.(2023)Asai, Wu, Wang, Sil, and Hajishirzi}]{selfrag}
Akari Asai, Zeqiu Wu, Yizhong Wang, Avirup Sil, and Hannaneh Hajishirzi. 2023.
\newblock \href {https://doi.org/10.48550/ARXIV.2310.11511} {Self-rag: Learning
  to retrieve, generate, and critique through self-reflection}.
\newblock \emph{CoRR}, abs/2310.11511.

\bibitem[{Bai et~al.(2022)Bai, Jones, Ndousse, Askell, Chen, DasSarma, Drain,
  Fort, Ganguli, Henighan, Joseph, Kadavath, Kernion, Conerly, Showk, Elhage,
  Hatfield{-}Dodds, Hernandez, Hume, Johnston, Kravec, Lovitt, Nanda, Olsson,
  Amodei, Brown, Clark, McCandlish, Olah, Mann, and Kaplan}]{HHData}
Yuntao Bai, Andy Jones, Kamal Ndousse, Amanda Askell, Anna Chen, Nova DasSarma,
  Dawn Drain, Stanislav Fort, Deep Ganguli, Tom Henighan, Nicholas Joseph,
  Saurav Kadavath, Jackson Kernion, Tom Conerly, Sheer~El Showk, Nelson Elhage,
  Zac Hatfield{-}Dodds, Danny Hernandez, Tristan Hume, Scott Johnston, Shauna
  Kravec, Liane Lovitt, Neel Nanda, Catherine Olsson, Dario Amodei, Tom~B.
  Brown, Jack Clark, Sam McCandlish, Chris Olah, Benjamin Mann, and Jared
  Kaplan. 2022.
\newblock \href {https://doi.org/10.48550/ARXIV.2204.05862} {Training a helpful
  and harmless assistant with reinforcement learning from human feedback}.
\newblock \emph{CoRR}, abs/2204.05862.

\bibitem[{Brown et~al.(2020)Brown, Mann, Ryder, Subbiah, Kaplan, Dhariwal,
  Neelakantan, Shyam, Sastry, Askell, Agarwal, Herbert{-}Voss, Krueger,
  Henighan, Child, Ramesh, Ziegler, Wu, Winter, Hesse, Chen, Sigler, Litwin,
  Gray, Chess, Clark, Berner, McCandlish, Radford, Sutskever, and
  Amodei}]{DBLP:conf/nips/BrownMRSKDNSSAA20}
Tom~B. Brown, Benjamin Mann, Nick Ryder, Melanie Subbiah, Jared Kaplan,
  Prafulla Dhariwal, Arvind Neelakantan, Pranav Shyam, Girish Sastry, Amanda
  Askell, Sandhini Agarwal, Ariel Herbert{-}Voss, Gretchen Krueger, Tom
  Henighan, Rewon Child, Aditya Ramesh, Daniel~M. Ziegler, Jeffrey Wu, Clemens
  Winter, Christopher Hesse, Mark Chen, Eric Sigler, Mateusz Litwin, Scott
  Gray, Benjamin Chess, Jack Clark, Christopher Berner, Sam McCandlish, Alec
  Radford, Ilya Sutskever, and Dario Amodei. 2020.
\newblock \href
  {https://proceedings.neurips.cc/paper/2020/hash/1457c0d6bfcb4967418bfb8ac142f64a-Abstract.html}
  {Language models are few-shot learners}.
\newblock In \emph{Advances in Neural Information Processing Systems 33: Annual
  Conference on Neural Information Processing Systems 2020, NeurIPS 2020,
  December 6-12, 2020, virtual}.

\bibitem[{Cohen et~al.(2009)Cohen, Huang, Chen, Benesty, Benesty, Chen, Huang,
  and Cohen}]{cohen2009pearson}
Israel Cohen, Yiteng Huang, Jingdong Chen, Jacob Benesty, Jacob Benesty,
  Jingdong Chen, Yiteng Huang, and Israel Cohen. 2009.
\newblock Pearson correlation coefficient.
\newblock \emph{Noise reduction in speech processing}, pages 1--4.

\bibitem[{Cohen et~al.(2023)Cohen, Hamri, Geva, and Globerson}]{DetectLMError}
Roi Cohen, May Hamri, Mor Geva, and Amir Globerson. 2023.
\newblock \href {https://doi.org/10.48550/ARXIV.2305.13281} {{LM} vs {LM:}
  detecting factual errors via cross examination}.
\newblock \emph{CoRR}, abs/2305.13281.

\bibitem[{Ding et~al.(2024)Ding, Pang, Wei, Shen, and
  Cheng}]{Ding_RetrieveOnlyWhen}
Hanxing Ding, Liang Pang, Zihao Wei, Huawei Shen, and Xueqi Cheng. 2024.
\newblock \href {https://doi.org/10.48550/ARXIV.2402.10612} {Retrieve only when
  it needs: Adaptive retrieval augmentation for hallucination mitigation in
  large language models}.
\newblock \emph{CoRR}, abs/2402.10612.

\bibitem[{Du et~al.(2023)Du, Li, Torralba, Tenenbaum, and
  Mordatch}]{ImprovingFactuality}
Yilun Du, Shuang Li, Antonio Torralba, Joshua~B. Tenenbaum, and Igor Mordatch.
  2023.
\newblock \href {https://doi.org/10.48550/ARXIV.2305.14325} {Improving
  factuality and reasoning in language models through multiagent debate}.
\newblock \emph{CoRR}, abs/2305.14325.

\bibitem[{Gawlikowski et~al.(2021)Gawlikowski, Tassi, Ali, Lee, Humt, Feng,
  Kruspe, Triebel, Jung, Roscher et~al.}]{uncertaintysurvey}
Jakob Gawlikowski, Cedrique Rovile~Njieutcheu Tassi, Mohsin Ali, Jongseok Lee,
  Matthias Humt, Jianxiang Feng, Anna Kruspe, Rudolph Triebel, Peter Jung,
  Ribana Roscher, et~al. 2021.
\newblock A survey of uncertainty in deep neural networks.
\newblock \emph{arXiv preprint arXiv:2107.03342}.

\bibitem[{Geng et~al.(2023)Geng, Cai, Wang, Koeppl, Nakov, and
  Gurevych}]{ASurveyofLMCalibration}
Jiahui Geng, Fengyu Cai, Yuxia Wang, Heinz Koeppl, Preslav Nakov, and Iryna
  Gurevych. 2023.
\newblock \href {https://doi.org/10.48550/ARXIV.2311.08298} {A survey of
  language model confidence estimation and calibration}.
\newblock \emph{CoRR}, abs/2311.08298.

\bibitem[{Guo et~al.(2017)Guo, Pleiss, Sun, and
  Weinberger}]{OnCalibrationofModernNN}
Chuan Guo, Geoff Pleiss, Yu~Sun, and Kilian~Q. Weinberger. 2017.
\newblock \href {http://proceedings.mlr.press/v70/guo17a.html} {On calibration
  of modern neural networks}.
\newblock In \emph{Proceedings of the 34th International Conference on Machine
  Learning, {ICML} 2017, Sydney, NSW, Australia, 6-11 August 2017}, volume~70
  of \emph{Proceedings of Machine Learning Research}, pages 1321--1330. {PMLR}.

\bibitem[{Guo et~al.(2021)Guo, Tan, Liu, Xing, and Hu}]{TextGeneration}
Han Guo, Bowen Tan, Zhengzhong Liu, Eric~P. Xing, and Zhiting Hu. 2021.
\newblock \href {http://arxiv.org/abs/2106.07704} {Text generation with
  efficient (soft) q-learning}.
\newblock \emph{CoRR}, abs/2106.07704.

\bibitem[{Ji et~al.(2023)Ji, Lee, Frieske, Yu, Su, Xu, Ishii, Bang, Madotto,
  and Fung}]{SurveyHallucination}
Ziwei Ji, Nayeon Lee, Rita Frieske, Tiezheng Yu, Dan Su, Yan Xu, Etsuko Ishii,
  Yejin Bang, Andrea Madotto, and Pascale Fung. 2023.
\newblock \href {https://doi.org/10.1145/3571730} {Survey of hallucination in
  natural language generation}.
\newblock \emph{{ACM} Comput. Surv.}, 55(12):248:1--248:38.

\bibitem[{Jiang et~al.(2023)Jiang, Sablayrolles, Mensch, Bamford, Chaplot,
  de~Las~Casas, Bressand, Lengyel, Lample, Saulnier, Lavaud, Lachaux, Stock,
  Scao, Lavril, Wang, Lacroix, and Sayed}]{Mistral}
Albert~Q. Jiang, Alexandre Sablayrolles, Arthur Mensch, Chris Bamford,
  Devendra~Singh Chaplot, Diego de~Las~Casas, Florian Bressand, Gianna Lengyel,
  Guillaume Lample, Lucile Saulnier, L{\'{e}}lio~Renard Lavaud, Marie{-}Anne
  Lachaux, Pierre Stock, Teven~Le Scao, Thibaut Lavril, Thomas Wang,
  Timoth{\'{e}}e Lacroix, and William~El Sayed. 2023.
\newblock \href {https://doi.org/10.48550/ARXIV.2310.06825} {Mistral 7b}.
\newblock \emph{CoRR}, abs/2310.06825.

\bibitem[{Kwiatkowski et~al.(2019)Kwiatkowski, Palomaki, Redfield, Collins,
  Parikh, Alberti, Epstein, Polosukhin, Devlin, Lee, Toutanova, Jones, Kelcey,
  Chang, Dai, Uszkoreit, Le, and Petrov}]{NaturalQ}
Tom Kwiatkowski, Jennimaria Palomaki, Olivia Redfield, Michael Collins,
  Ankur~P. Parikh, Chris Alberti, Danielle Epstein, Illia Polosukhin, Jacob
  Devlin, Kenton Lee, Kristina Toutanova, Llion Jones, Matthew Kelcey,
  Ming{-}Wei Chang, Andrew~M. Dai, Jakob Uszkoreit, Quoc Le, and Slav Petrov.
  2019.
\newblock \href {https://doi.org/10.1162/TACL\_A\_00276} {Natural questions: a
  benchmark for question answering research}.
\newblock \emph{Trans. Assoc. Comput. Linguistics}, 7:452--466.

\bibitem[{Lin et~al.(2022{\natexlab{a}})Lin, Hilton, and
  Evans}]{lin2022teaching}
Stephanie Lin, Jacob Hilton, and Owain Evans. 2022{\natexlab{a}}.
\newblock Teaching models to express their uncertainty in words.
\newblock \emph{arXiv preprint arXiv:2205.14334}.

\bibitem[{Lin et~al.(2022{\natexlab{b}})Lin, Hilton, and Evans}]{Truthfulqa}
Stephanie Lin, Jacob Hilton, and Owain Evans. 2022{\natexlab{b}}.
\newblock \href {https://doi.org/10.18653/V1/2022.ACL-LONG.229} {Truthfulqa:
  Measuring how models mimic human falsehoods}.
\newblock In \emph{Proceedings of the 60th Annual Meeting of the Association
  for Computational Linguistics (Volume 1: Long Papers), {ACL} 2022, Dublin,
  Ireland, May 22-27, 2022}, pages 3214--3252. Association for Computational
  Linguistics.

\bibitem[{Liu et~al.(2023)Liu, Yao, Ton, Zhang, Guo, Cheng, Klochkov, Taufiq,
  and Li}]{DBLP:journals/corr/abs-2308-05374}
Yang Liu, Yuanshun Yao, Jean{-}Francois Ton, Xiaoying Zhang, Ruocheng Guo, Hao
  Cheng, Yegor Klochkov, Muhammad~Faaiz Taufiq, and Hang Li. 2023.
\newblock \href {https://doi.org/10.48550/ARXIV.2308.05374} {Trustworthy llms:
  a survey and guideline for evaluating large language models' alignment}.
\newblock \emph{CoRR}, abs/2308.05374.

\bibitem[{Mielke et~al.(2022)Mielke, Szlam, Dinan, and
  Boureau}]{reducingoverconfidence}
Sabrina~J Mielke, Arthur Szlam, Emily Dinan, and Y-Lan Boureau. 2022.
\newblock Reducing conversational agents’ overconfidence through linguistic
  calibration.
\newblock \emph{Transactions of the Association for Computational Linguistics},
  10:857--872.

\bibitem[{Minderer et~al.(2021)Minderer, Djolonga, Romijnders, Hubis, Zhai,
  Houlsby, Tran, and Lucic}]{RevisitTheCalibration}
Matthias Minderer, Josip Djolonga, Rob Romijnders, Frances Hubis, Xiaohua Zhai,
  Neil Houlsby, Dustin Tran, and Mario Lucic. 2021.
\newblock \href
  {https://proceedings.neurips.cc/paper/2021/hash/8420d359404024567b5aefda1231af24-Abstract.html}
  {Revisiting the calibration of modern neural networks}.
\newblock In \emph{Advances in Neural Information Processing Systems 34: Annual
  Conference on Neural Information Processing Systems 2021, NeurIPS 2021,
  December 6-14, 2021, virtual}, pages 15682--15694.

\bibitem[{OpenAI(2023)}]{gpt4}
OpenAI. 2023.
\newblock \href {https://doi.org/10.48550/ARXIV.2303.08774} {{GPT-4} technical
  report}.
\newblock \emph{CoRR}, abs/2303.08774.

\bibitem[{Ramamurthy et~al.(2023)Ramamurthy, Ammanabrolu, Brantley, Hessel,
  Sifa, Bauckhage, Hajishirzi, and Choi}]{Ramamurthy2022IsRL}
Rajkumar Ramamurthy, Prithviraj Ammanabrolu, Kianté Brantley, Jack Hessel,
  Rafet Sifa, Christian Bauckhage, Hannaneh Hajishirzi, and Yejin Choi. 2023.
\newblock \href {https://arxiv.org/abs/2210.01241} {Is reinforcement learning
  (not) for natural language processing: Benchmarks, baselines, and building
  blocks for natural language policy optimization}.
\newblock In \emph{International Conference on Learning Representations
  (ICLR)}.

\bibitem[{Schulman et~al.(2017)Schulman, Wolski, Dhariwal, Radford, and
  Klimov}]{PPO}
John Schulman, Filip Wolski, Prafulla Dhariwal, Alec Radford, and Oleg Klimov.
  2017.
\newblock \href {http://arxiv.org/abs/1707.06347} {Proximal policy optimization
  algorithms}.
\newblock \emph{CoRR}, abs/1707.06347.

\bibitem[{Sedgwick(2014)}]{sedgwick2014spearman}
Philip Sedgwick. 2014.
\newblock Spearman’s rank correlation coefficient.
\newblock \emph{Bmj}, 349.

\bibitem[{Tian et~al.(2023)Tian, Mitchell, Zhou, Sharma, Rafailov, Yao, Finn,
  and Manning}]{JustAsk}
Katherine Tian, Eric Mitchell, Allan Zhou, Archit Sharma, Rafael Rafailov,
  Huaxiu Yao, Chelsea Finn, and Christopher~D. Manning. 2023.
\newblock \href {https://doi.org/10.48550/ARXIV.2305.14975} {Just ask for
  calibration: Strategies for eliciting calibrated confidence scores from
  language models fine-tuned with human feedback}.
\newblock \emph{CoRR}, abs/2305.14975.

\bibitem[{Touvron et~al.(2023)Touvron, Martin, Stone, Albert, Almahairi,
  Babaei, Bashlykov, Batra, Bhargava, Bhosale, Bikel, Blecher, Canton{-}Ferrer,
  Chen, Cucurull, Esiobu, Fernandes, Fu, Fu, Fuller, Gao, Goswami, Goyal,
  Hartshorn, Hosseini, Hou, Inan, Kardas, Kerkez, Khabsa, Kloumann, Korenev,
  Koura, Lachaux, Lavril, Lee, Liskovich, Lu, Mao, Martinet, Mihaylov, Mishra,
  Molybog, Nie, Poulton, Reizenstein, Rungta, Saladi, Schelten, Silva, Smith,
  Subramanian, Tan, Tang, Taylor, Williams, Kuan, Xu, Yan, Zarov, Zhang, Fan,
  Kambadur, Narang, Rodriguez, Stojnic, Edunov, and Scialom}]{LLAMA}
Hugo Touvron, Louis Martin, Kevin Stone, Peter Albert, Amjad Almahairi, Yasmine
  Babaei, Nikolay Bashlykov, Soumya Batra, Prajjwal Bhargava, Shruti Bhosale,
  Dan Bikel, Lukas Blecher, Cristian Canton{-}Ferrer, Moya Chen, Guillem
  Cucurull, David Esiobu, Jude Fernandes, Jeremy Fu, Wenyin Fu, Brian Fuller,
  Cynthia Gao, Vedanuj Goswami, Naman Goyal, Anthony Hartshorn, Saghar
  Hosseini, Rui Hou, Hakan Inan, Marcin Kardas, Viktor Kerkez, Madian Khabsa,
  Isabel Kloumann, Artem Korenev, Punit~Singh Koura, Marie{-}Anne Lachaux,
  Thibaut Lavril, Jenya Lee, Diana Liskovich, Yinghai Lu, Yuning Mao, Xavier
  Martinet, Todor Mihaylov, Pushkar Mishra, Igor Molybog, Yixin Nie, Andrew
  Poulton, Jeremy Reizenstein, Rashi Rungta, Kalyan Saladi, Alan Schelten, Ruan
  Silva, Eric~Michael Smith, Ranjan Subramanian, Xiaoqing~Ellen Tan, Binh Tang,
  Ross Taylor, Adina Williams, Jian~Xiang Kuan, Puxin Xu, Zheng Yan, Iliyan
  Zarov, Yuchen Zhang, Angela Fan, Melanie Kambadur, Sharan Narang,
  Aur{\'{e}}lien Rodriguez, Robert Stojnic, Sergey Edunov, and Thomas Scialom.
  2023.
\newblock \href {https://doi.org/10.48550/ARXIV.2307.09288} {Llama 2: Open
  foundation and fine-tuned chat models}.
\newblock \emph{CoRR}, abs/2307.09288.

\bibitem[{Tunstall et~al.(2023)Tunstall, Beeching, Lambert, Rajani, Rasul,
  Belkada, Huang, von Werra, Fourrier, Habib, Sarrazin, Sanseviero, Rush, and
  Wolf}]{Zephyr}
Lewis Tunstall, Edward Beeching, Nathan Lambert, Nazneen Rajani, Kashif Rasul,
  Younes Belkada, Shengyi Huang, Leandro von Werra, Cl{\'{e}}mentine Fourrier,
  Nathan Habib, Nathan Sarrazin, Omar Sanseviero, Alexander~M. Rush, and Thomas
  Wolf. 2023.
\newblock \href {https://doi.org/10.48550/ARXIV.2310.16944} {Zephyr: Direct
  distillation of {LM} alignment}.
\newblock \emph{CoRR}, abs/2310.16944.

\bibitem[{Wei et~al.(2022)Wei, Wang, Schuurmans, Bosma, Chi, Le, and
  Zhou}]{COT}
Jason Wei, Xuezhi Wang, Dale Schuurmans, Maarten Bosma, Ed~Chi, Quoc Le, and
  Denny Zhou. 2022.
\newblock Chain of thought prompting elicits reasoning in large language
  models.
\newblock \emph{arXiv preprint arXiv:2201.11903}.

\bibitem[{Wu et~al.(2023)Wu, Hu, Shi, Dziri, Suhr, Ammanabrolu, Smith,
  Ostendorf, and Hajishirzi}]{Finegrained}
Zeqiu Wu, Yushi Hu, Weijia Shi, Nouha Dziri, Alane Suhr, Prithviraj
  Ammanabrolu, Noah~A. Smith, Mari Ostendorf, and Hannaneh Hajishirzi. 2023.
\newblock \href {https://doi.org/10.48550/ARXIV.2306.01693} {Fine-grained human
  feedback gives better rewards for language model training}.
\newblock \emph{CoRR}, abs/2306.01693.

\bibitem[{Xiong et~al.(2023)Xiong, Hu, Lu, Li, Fu, He, and
  Hooi}]{CanLLMExpress}
Miao Xiong, Zhiyuan Hu, Xinyang Lu, Yifei Li, Jie Fu, Junxian He, and Bryan
  Hooi. 2023.
\newblock \href {https://doi.org/10.48550/ARXIV.2306.13063} {Can llms express
  their uncertainty? an empirical evaluation of confidence elicitation in
  llms}.
\newblock \emph{CoRR}, abs/2306.13063.

\bibitem[{Zhang et~al.(2023{\natexlab{a}})Zhang, Cui, Bi, and
  Shi}]{DBLP:journals/corr/abs-2312-15710}
Yue Zhang, Leyang Cui, Wei Bi, and Shuming Shi. 2023{\natexlab{a}}.
\newblock Alleviating hallucinations of large language models through induced
  hallucinations.
\newblock \emph{CoRR}, abs/2312.15710.

\bibitem[{Zhang et~al.(2023{\natexlab{b}})Zhang, Li, Cui, Cai, Liu, Fu, Huang,
  Zhao, Zhang, Chen, Wang, Luu, Bi, Shi, and Shi}]{SurveySiren}
Yue Zhang, Yafu Li, Leyang Cui, Deng Cai, Lemao Liu, Tingchen Fu, Xinting
  Huang, Enbo Zhao, Yu~Zhang, Yulong Chen, Longyue Wang, Anh~Tuan Luu, Wei Bi,
  Freda Shi, and Shuming Shi. 2023{\natexlab{b}}.
\newblock \href {https://doi.org/10.48550/arXiv.2309.01219} {Siren's song in
  the {AI} ocean: {A} survey on hallucination in large language models}.
\newblock \emph{CoRR}, abs/2309.01219.

\bibitem[{Zheng et~al.(2023)Zheng, Dou, Gao, Hua, Shen, Wang, Liu, Jin, Liu,
  Zhou, Xiong, Chen, Xi, Xu, Lai, Zhu, Chang, Yin, Weng, Cheng, Huang, Sun,
  Yan, Gui, Zhang, Qiu, and Huang}]{SecretPPO}
Rui Zheng, Shihan Dou, Songyang Gao, Yuan Hua, Wei Shen, Binghai Wang, Yan Liu,
  Senjie Jin, Qin Liu, Yuhao Zhou, Limao Xiong, Lu~Chen, Zhiheng Xi, Nuo Xu,
  Wenbin Lai, Minghao Zhu, Cheng Chang, Zhangyue Yin, Rongxiang Weng, Wensen
  Cheng, Haoran Huang, Tianxiang Sun, Hang Yan, Tao Gui, Qi~Zhang, Xipeng Qiu,
  and Xuanjing Huang. 2023.
\newblock \href {https://doi.org/10.48550/ARXIV.2307.04964} {Secrets of {RLHF}
  in large language models part {I:} {PPO}}.
\newblock \emph{CoRR}, abs/2307.04964.

\bibitem[{Zhou et~al.(2023)Zhou, Jurafsky, and Hashimoto}]{navigating}
Kaitlyn Zhou, Dan Jurafsky, and Tatsunori Hashimoto. 2023.
\newblock Navigating the grey area: Expressions of overconfidence and
  uncertainty in language models.
\newblock \emph{arXiv preprint arXiv:2302.13439}.

\bibitem[{Ziegler et~al.(2019)Ziegler, Stiennon, Wu, Brown, Radford, Amodei,
  Christiano, and Irving}]{FTHP}
Daniel~M. Ziegler, Nisan Stiennon, Jeffrey Wu, Tom~B. Brown, Alec Radford,
  Dario Amodei, Paul~F. Christiano, and Geoffrey Irving. 2019.
\newblock \href {http://arxiv.org/abs/1909.08593} {Fine-tuning language models
  from human preferences}.
\newblock \emph{CoRR}, abs/1909.08593.

\end{thebibliography}

\appendix

\appendix

\clearpage
\section{Text Generation Environment}
\label{apd:rl_environment}
We provide a reinforcement learning environment for text generation in LLM.

Our study concentrates on text-generation tasks. For each task, we are provided with a collection of input prompts, denoted as $D=\{x^n\}_{n=1}^N$. We follow the framework of~\cite{Ramamurthy2022IsRL,Finegrained}) to model language generation as a Markov Decision Process (MDP), represented by the tuple $\langle \mathcal{S}, \mathcal{A}, \mathcal{R}, P, \gamma, T_{max} \rangle$, where $\mathcal{V}$ is a finite vocabulary set.

In this MDP, an episode commences with a randomly chosen prompt $x=(x_1, x_2, \dots, x_l)$, where each $x_i$ is an element of $\mathcal{V}$. The episode concludes either when the sequence generation surpasses the maximum time step $T_{max}$ or when an end-of-sequence token is produced. The state space is denoted by $\mathcal{S}$, with the initial state $s_0=(x_1, x_2, \dots, x_l)$ belonging to $\mathcal{S}$. An action $a_t \in \mathcal{A}$, which is a token generated by the policy language model $P_\theta$ at time $t$, is selected from $\mathcal{V}$, with $a_0$ signifying the beginning of the sequence. The transition function $P$ extends the current state $s_t$ by appending the action $a_t$, resulting in the sequence $(x_1, x_2, \dots, x_l, a_0, a_1, \dots, a_{t-1})$. This iterative process is maintained until the final time step $T\le T_{max}$, yielding the output sequence $y=(a_1, \dots, a_T)$.

The reward function $\mathcal{R}$ assigns real-valued rewards based on the criteria detailed in Section 5, providing feedback at each step and upon reaching time $T$. The policy model $P_\theta$ may be initialized using a pre-trained language model and potentially further refined through supervised fine-tuning using task-specific examples. The specifics of the reward function are elaborated upon later in the text.

\section{Prompt}
\label{apd:prompt}
\subsection{Prompts for GPT evaluation}
The prompt for GPT evaluation is:

\texttt{
Determine whether the given Answer and Reference for a specified Question are semantically equivalent. If the Answer matches or aligns with the Reference's meaning, or if the Answer is included in the Reference, rate it as 1.0. If not, rate it as 0.0. Provide only 0.0 or 1.0, with no other choice or explanation.
\textbackslash n\#\#\# Question: \$\{QUESTION\}.
\textbackslash n\#\#\# Answer: \$\{OUTPUT\_ANSWER\}.
\textbackslash n\#\#\# Reference: \$\{GOLDEN\_ANSWER\}.
\textbackslash n\#\#\# Rating:
}

\subsection{Prompts for Inducing the Verbalized Confidence}
\paragraph{Vanilla \& Ours}
\texttt{
Please provide the answer and the Confidence (0.0 to 1.0) that the answer is truthful and informative for the Question.\textbackslash n\#\#\# Question: who played will on as the world turns?\textbackslash n \textbackslash n \#\#\# Answer: Jesse Soffer.\textbackslash n\#\#\# Confidence: 0.9.\textbackslash n\#\#\# Question: how many episodes in season 2 breaking bad?\textbackslash n\#\#\# Answer: 5\textbackslash n\#\#\# Confidence: 0.2.\textbackslash n\#\#\# Question: 
}
Note that to ensure a fair comparison, our CONQORD also adopts the above prompt to induce confidence.
\paragraph{Top-k}
\texttt{
Provide your 2 best answers and the Confidence that each is correct (0.0 to 1.0) for the following question. Give ONLY the task output description of your guesses and probabilities, no other words or explanations. The format is:
\#\#\# Question: \textbackslash n\#\#\# Answer-1: Yes. Bruno Richard Hauptmann was sentenced to death for the kidnapping.\textbackslash n\#\#\# Confidence-1: 0.9.\textbackslash n\#\#\# Answer-2: No.\textbackslash n\#\#\# Confidence-2: 0.0.\textbackslash n\textbackslash n
\#\#\# Question:
}
\paragraph{CoT+Agg}
\texttt{
Please provide the answer and the Confidence (0.0 to 1.0) that the answer is truthful and informative for the Question. Let’s think it step by step.\textbackslash n\textbackslash n
\#\#\# Question: Was the Lindbergh kidnapping ever solved?\textbackslash n\#\#\# Explanation: Bruno Richard Hauptmann was sentenced to death for the kidnapping. \textbackslash n\#\#\# Answer: Yes.\textbackslash n\#\#\# Confidence: 0.9.\textbackslash n\textbackslash n
\#\#\# Question:
}
\subsection{Prompts for RAG }
\texttt{
Given the following question and the retrieved evidence, please generate a truthful and informative answer. Only answer, no explanations.\textbackslash n\textbackslash n\textbackslash n\#\#\# Question: {question}\textbackslash n\textbackslash n\textbackslash n\#\#\# Retrieved Evidences: {rag}\textbackslash n\textbackslash n\textbackslash n\#\#\# Answer:
}
\clearpage
\section{Examples of Constructed Tuple in PreApproach}
\label{apd:construct_data}
Chosen:
\begin{itemize}
	\item 
	\texttt{\#\#\# Question: \$\{QUESTION\}. \\
\#\#\# Answer: \$\{GOOD\_ANSWER\}. \\
\#\#\# Confidence: 0.9.}
\item 
\texttt{\#\#\# Question: \$\{QUESTION\}.  \\
\#\#\# Answer: \$\{BAD\_ANSWER\}. \\
\#\#\# Confidence: 0.1. }
\end{itemize}

\noindent Rejected:
\begin{itemize}
	\item 
	\texttt{\#\#\# Question: \$\{QUESTION\}. \\
\#\#\# Answer: \$\{GOOD\_ANSWER\}. \\
\#\#\# Confidence: 0.1.}
\item 
\texttt{\#\#\# Question: \$\{QUESTION\}. \\
\#\#\# Answer: \$\{BAD\_ANSWER\}. \\
\#\#\# Confidence: 0.9. }
\end{itemize}

\label{sec:appendix}

\end{document}